%% file: cl_workshop_onlinegp.tex
\documentclass{article}
\usepackage[latin1]{inputenc}
\usepackage{amsmath}
\usepackage{amsfonts}
\usepackage{amssymb}
\usepackage{hyperref}
\usepackage{graphicx}
\usepackage[accepted]{icml2020}
\usepackage[acronym]{glossaries}
\usepackage{subcaption}
\input{commands}
\usepackage{pifont}

\newcommand{\kDPP}{k-DPP}
\newcommand{\namealg}{OIPS}
\newcommand{\setZ}{Z}
\icmltitlerunning{Online Inducing Points Selection for Gaussian Processes}
\begin{document}

\twocolumn[
\icmltitle{Adaptive Inducing Points Selection for Gaussian Processes}




\begin{icmlauthorlist}
	\icmlauthor{Th\'eo Galy-Fajou}{tu}
	\icmlauthor{Manfred Opper}{tu}
\end{icmlauthorlist}

\icmlaffiliation{tu}{TU Berlin}

\icmlcorrespondingauthor{Th\'eo Galy-Fajou}{galyfajou@tu-berlin.de}

\begin{center} $^1$ Technical University of Berlin\end{center}
\icmlkeywords{Machine Learning, ICML, Gaussian Process, Streaming, Online}

\vskip 0.3in
]
\begin{abstract}
Gaussian Processes (\textbf{GPs}) are flexible non-parametric models with strong probabilistic interpretation.
While being a standard choice for performing inference on time series, GPs have little techniques to work in a streaming setting.
\cite{bui2017streaming} developed an efficient variational approach to train online GPs by using sparsity techniques:
The whole set of observations is approximated by a smaller set of inducing points (\textbf{IPs}) and moved around with new data.
Both the number and the locations of the IPs will affect greatly the performance of the algorithm.
In addition to optimizing their locations we propose to adaptively add new points, based on the properties of the GP and the structure of the data.
\end{abstract}
\vspace{-1cm}
\section{Introduction}

Gaussian Processes (\textbf{GPs}) are flexible non-parametric models with strong probabilistic interpretation.
They are particularly fitted for time-series \cite{roberts_gaussian_2013} but one of their biggest limitations is that they scale cubically with the number of points \cite{williams2006gaussian}.
\citet{quinonerocandelaunifying2005} introduced the notion of sparse GPs, models approximating the posterior by a smaller number $M$ of inducing points (\textbf{IPs}) and reducing the inference complexity from $\mathcal{O}(N^3)$ to $\mathcal{O}(M^3)$ where $M$ is the number of IPs. 
\citet{titsias2009variational} introduced them later in a variational setting, allowing to optimize their locations.
Based on this idea, \cite{bui2017streaming} introduced a variational streaming model relying on inducing points.
One of their algorithm's features is that hyper-parameters can be optimized and more specifically the number of inducing can vary between batches of data.
However in their work, the number of IPs is fixed and their locations are simply optimized against the variational bound of the marginal likelihood.
Having a fixed number of IPs limits the model's scope if the total data size is unknown.
A gradient based approach leads to two problems:\\
-~ IP's locations need to be optimized until convergence for every batch. Therefore  batches need to be sufficiently large to get a meaningful improvement.
If the new data comes in very far from the original positions of the IPs, the optimization will be extremely slow.\\
-~ The number of IPs being fixed, there is no way to know how many will be required to have a desired accuracy.
Finding the optimal number of IPs is also not an option as it is an ill-posed problem: the objective will only decrease with more IPs, i.e. the optimum is obtained when every data point is an IP.

\begin{figure}
	\begin{center}
	\includegraphics[width=\columnwidth]{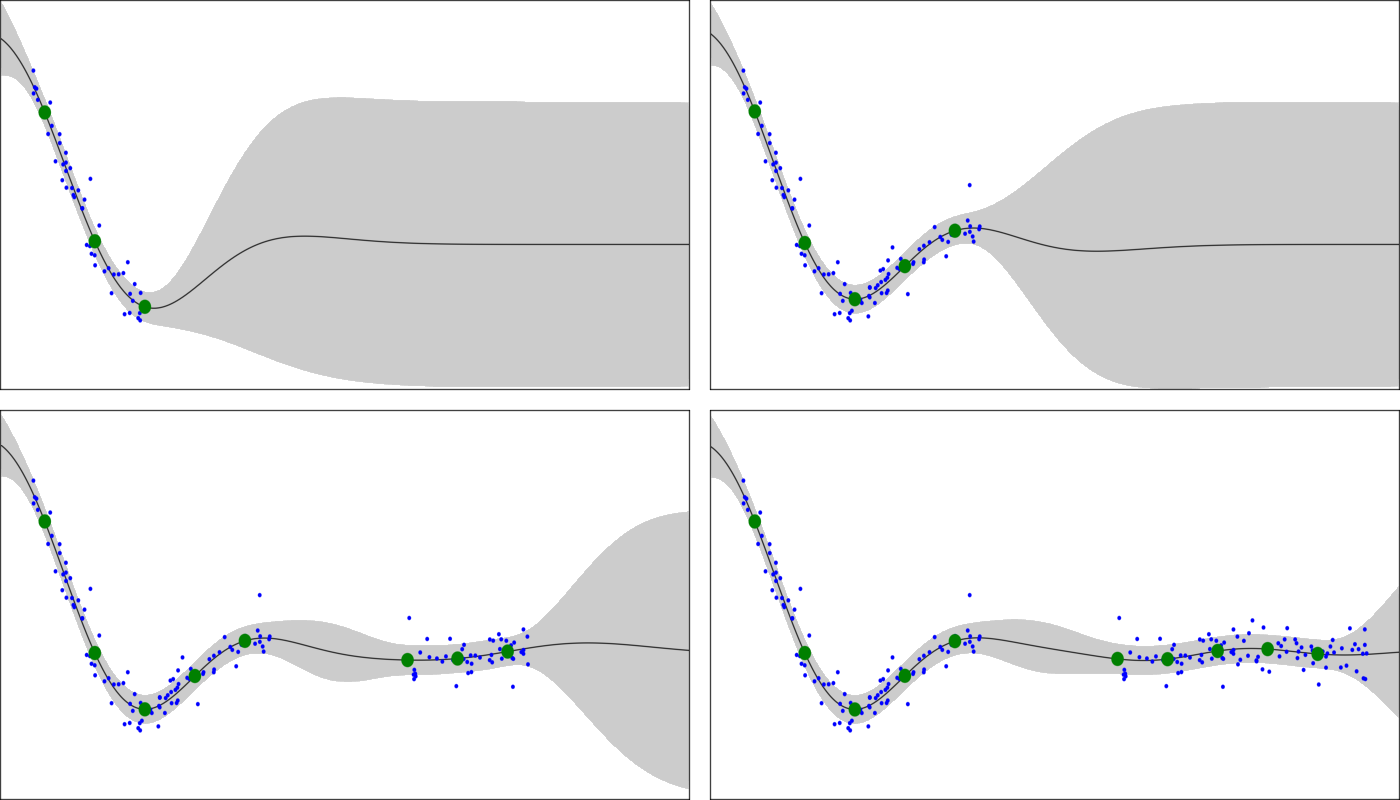}
	\end{center}
	\caption{Illustration of the inducing point selection process. Blue points represent inducing points, green points data and the orange line represent the mean of the prediction from the GP model surrounded by one standard error.
	The dashed represent the space covered by the existing IPs, only points seen outside those areas are selected as new IPs.}
\end{figure}
We propose a different approach to this problem with a simple algorithm, \textbf{Online Inducing Points Selection} (\textbf{\namealg}), requiring only one parameter to select automatically both the number of inducing points and their location.
\namealg~ naturally takes into account the structure of the data while the performance trade-off and the expected number of IPs can be inferred.

Our main contributions are as follow :\\
- We develop an efficient online algorithm to automatically select the number and location of inducing points for a streaming GP.\\
- We give theoretical guarantees on the expected number of inducing points and the performance of the GP.

In section \ref{sec:background} we present existing methods to select inducing points, as well as an online inference for GPs.
We present our algorithm and its theoretical guarantees in section \ref{sec:algorithm}.
We show our experiments in comparison with popular inducing points selection methods in section \ref{sec:experiments}.
Finally we summarize our findings and explore outlooks in section \ref{sec:conclusion}.

\section{Background}
\label{sec:background}
\subsection{Sparse Variational Gaussian Processes}

\paragraph{Gaussian Processes:} Given some training data $\mathcal{D} = \{X,\boldy\}$ where $X= \{x_i\}_{i=1}^N $ are the inputs $x_i\in\mathbb{R}^D$ and $\boldy= \{y_i\}_{i=1}^N$ are the labels, we want to compute the predictive distribution $p(y^*|D,x^*)$ for new inputs $x^*$.
In order to do this we try to find an optimal distribution over a latent function $f$.
We set the latent vector $\boldf$ as the realization of $f(X)$, where $f_i = f(x_i)$, and put a GP prior $\mathcal{GP}(\mu_0,k)$ on $\boldf$, with $\mu_0$ the prior mean (set to 0 without loss of generality) and $k$ a kernel function.
In this work we are going to use an isotropic squared exponential kernel (\textbf{SE kernel}) : ${k(x,x') = \exp(-||x-x'||^2/l^2)}$, but it is generally applicable to all translation-invariant kernels.
We then compute the posterior:
\begin{align}
p(\boldf|\mathcal{D}) = \frac{\prod_{i=1}^Np(y_i|f_i)p(\boldf)}{p(\mathcal{D})}
\end{align}
Where $p(\boldf) \sim \mathcal{N}(0,K_{XX})$ and $K_{XX}$ is the kernel matrix evaluated on $X$ (in later notation we use $K_X$ instead of $K_{XX}$).
For a Gaussian likelihood the posterior $p(\boldf|\mathcal{D})$ is known analytically in closed-form.
Prediction and inference have nonetheless a complexity of $\mathcal{O}(N^3)$

\paragraph{Sparse Variational Gaussian Processes:} When the likelihood is not Gaussian, there is no tractable solution for the posterior. 
One possible approximation is to use variational inference : a family of distributions over $\boldf$ is selected, e.g. the multivariate Gaussian $q(\boldf)=\mathcal{N}(\boldm,S)$, and one optimizes the variational parameters $\boldm$ and $S$ by minimizing the negative ELBO, a proxy for the KL divergence $\KL(q(\boldf)||p(\boldf|\mathcal{D}))$.
However the computational complexity still grows cubically with the number of samples, and is therefore inadequate to large datasets.

\citet{quinonerocandelaunifying2005} and \citet{titsias2009variational} introduced the notion of sparse variational GPs (\textbf{SVGP}).
One adds inducing variables $\boldu$ and their inducing locations $Z=\{Z_i\}_{i=1}^M$ to the model.
In this work we restrict $Z_i$ to be in the same domain as $X_i$ but inter-domain approaches also exist \cite{hensman2017variational}.
The relation between $\boldu$ and $\boldf$ is given by the distribution
$p(\boldf,\boldu) = p(\boldf|\boldu)p(\boldu)$ where
\begin{align}
p(\boldf|\boldu) = \mathcal{N}(\boldf|K_{XZ}K_{Z}^{-1}\boldu,\widetilde{K}), \, p(\boldu) = \mathcal{N}(0,K_Z)\label{eq:sparse}
\end{align}
where $\widetilde{K}=K_{X}-K_{XZ}K_{Z}^{-1}K_{ZX}$

Then we approximate $p(\boldf,\boldu)$ with the variational distribution $q(\boldf,\boldu) = p(\boldf|\boldu)q(\boldu)$ where $q(\boldu) =\mathcal{N}(\bmu,\Sigma)$ by optimizing $\KL(q(\boldf,\boldu)||p(\boldf,\boldu|\mathcal{D}))$.

Note that if the likelihood is Gaussian, the optimal variational parameters $\bmu^*$ and $\Sigma^*$ are known in closed-form.
The only parameters left to optimize are the kernel parameters as well as selecting the number and the location of the inducing variables.

\subsection{Inducing points selection methods}

\label{sec:othermethods}
\citet{titsias2009variational} initially proposed to select the points location via a greedy selection : 
A small batch of data is randomly sampled, each sample is successively tested by adding it to the set of inducing points and evaluating the improvement on the ELBO.
The sample bringing the best performance is added to the set of inducing points and the operation is repeated until the desired number of inducing points is reached. 
This greedy approach has the advantage of selecting a set which is already close to the optimum set but is extremely expensive and is not applicable to non-conjugate likelihoods as it relies on estimating the optimal bound.

The most popular approach currently is to use the $k$-means++ algorithm \cite{arthur2007k} and take the optimized clusters centers as inducing points locations.
The clustering nature of the algorithm allows to have good coverage of the whole dataset.
However the $k$-means algorithm have a complexity of $\mathcal{O}(NMDT)$ on the whole dataset where $T$ is the number of $k$-means iterations.
Another issue is that it might allocate multiple centers in a region of high density leading to very close inducing points and no significant performance improvement.
It is also not applicable online and does not solve the problem of choosing the number of inducing points.

Another classical approach is to simply take a grid. 
For example \citet{MorenoArtesAlvarez19} use a grid in an online setting by updating the bounds of a uniform grid.
Using a grid is unfortunately limited a small number of dimensions and does not take into account the structure of the data.

\vspace{-0.5cm}
\subsection{Online Variational Gaussian Process Learning}
\cite{bui2017streaming} developed a streaming algorithm for GPs (\textbf{SSVGP}) based the inducing points approach of \cite{titsias2009variational}.
The method consists in recursively optimizing the variational distribution $q_t(\boldu_t,\boldf)$ for each new batch of data $\mathcal{D}_t$ given the previous variational distribution $q_{t-1}(\boldu_{t-1},\boldf)$.
$q_t$ initially approximates the posterior : 
\begin{align}
 p(\boldu_t,\hspace{-0.05cm}\boldf|\mathcal{D}_{1:t})
= \frac{p(\mathcal{D}_{t}|\boldf)p(\mathcal{D}_{1:(t-1)}|\boldf)p(\boldu_t,\hspace{-0.05cm}\boldf|\theta_t)}{p(\mathcal{D}_{1:t})}\label{eq:contlearning}
\end{align}
where $\theta_t$ are the set of hyper-parameters.
Since $D_{1:(t-1)}$ is not accessible anymore, the likelihood on previously seen data is approximated using the previous variational approximation $q_{t-1}(\boldu_{t-1})$ and the previous hyper-parameters $\theta_{t-1}$:
\begin{align*}
p(\mathcal{D}_{1:(t-1)}|\boldf) \approx \frac{q_{t-1}(\boldu_{t-1})p(\mathcal{D}_{1:(t-1)})}{p(\boldu_{t-1}|\theta_{t-1})}.
\end{align*}
The distribution approximated by $q_t$ is in the end:
\begin{align}
\begin{split}
& q_t(\boldu_t,\hspace{-0.05cm}\boldf|\mathcal{D}_{1:t})
\approx\\ &\quad\frac{p(\mathcal{D}_{t}|\boldf)q_{t-1}(\boldu_{t-1})p(\boldu_t,\hspace{-0.05cm}\boldf|\theta_t)}{p(\boldu_{t-1}|\theta_{t-1})}\frac{p(\mathcal{D}_{1:(t-1)})}{p(\mathcal{D}_{1:t})}
\end{split}\label{eq:contlearningfull}
\end{align}

The optimization of the (bound on the) KL divergence between the two distributions for each new batch will preserve the information of $\mathcal{D}_{1:(t-1)}$ via $q_{t-1}$ and ensure a smooth transition of the hyper-parameters, including the number of inducing points.
 We give all technical details including the hyper-parameter derivatives and the ELBO in full form in appendix \ref{appendix:elbo}.
\vspace{-0.5cm}
\section{Algorithm}
\label{sec:algorithm}
The idea of our algorithm is that to give a good approximation, a large majority of the samples should be "close" (in the reproducing kernel Hilbert space (RKHS)) to the set $\setZ$ of IPs locations.
Additionally, $\setZ$ should be as diverse as possible, since IP degeneracy will not improve the approximation.
This intuition is supported by previous works:\\
- \citet{bauer2016understanding} showed that the most substantial improvement obtained by adding a new inducing point was through the reduction of the uncertainty of $q(\boldf)$, which decreases quadratically with $K_{X\setZ}$.\\
- \citet{burt2019rates} showed that the quality of the approximation made with inducing points is bounded by the norm of ${Q_{X}=K_{X}-K_{X\setZ}K_{\setZ}^{-1}K_{\setZ X}}$.\\
Therefore by ensuring that $K_{X\setZ}$ and $|K_{\setZ}|$ are sufficiently large, we can expect an improvement on the approximation of the non-sparse problem.


\vspace{-0.2cm}

\subsection{Adding New Inducing Points}

A simple yet efficient strategy is to verify that for each new data point $x$ seen during training, there exists a close inducing point.
We first compute $K_{x\setZ} = [k(x,Z_1),\ldots, k(x,Z_M)]$.
If the maximum value of $K_{x\setZ}$ is smaller than a threshold parameter $\rho$, the sample is added to the set of IPs $\setZ$. If not, the algorithm passes on to the next sample.
We summarize all steps in Algorithm \ref{alg:add}.

The streaming nature of the algorithm makes it perfectly suited for an online learning setting :
it needs to see samples only once, whereas other algorithms like $k$-means need to parse all the data multiple times before converging.
It is fully deterministic for a given sequence of samples and therefore convergence guarantees are given under some conditions.
This approach was previously explored in a different context by \citet{csato2002sparse}, but was limited to small datasets.
\begin{algorithm}[t]
	\caption{Online Inducing Point Selection (\textbf{OIPS})}
	\label{alg:add}
	\begin{algorithmic}
	\STATE {\bfseries Input:} sample $x$, set of inducing points $\setZ = \{Z_j\}_{j=1}^M$,\\ acceptance threshold $0<\rho < 1$, kernel function $k$
	\STATE{$d \leftarrow \max_j(k(x,Z_j))$}
	\IF{$d < \rho$}
		\STATE{$\{Z_j\} \leftarrow \{Z_j\}\bigcup x$\\ $M\leftarrow M+1$}
	\ENDIF
	\STATE{\bfseries return} {$\{Z_j\}$}
	\end{algorithmic}
\end{algorithm}

The extra cost of the algorithm is virtually free since $K_{XZ}$ needs to be computed for the variational updates of the model.

One of our claims is that our algorithm is model and data agnostic.
The reason is that as kernel hyper-parameters are optimized, the acceptance condition changes as well

Note that this method can be interpreted as a half-greedy approach of a sequential sampling of a determinantal point process \cite{Kulesza2012}.
In appendix \ref{appendix:pdfdpp}, we show that for the same number of points, the probability of our selected set is higher than the one of a \kDPP.
\vspace{-0.3cm}

\subsection{Theoretical guarantees}

The final size of $\setZ$ is depending on many factors: the selected threshold $\rho$, the chosen kernel, the structure of the data (distribution, sparsity, etc) and the number of points seen.
However by having some weak assumptions on the data we can prove a bound on the expected number of inducing points as well as on the quality of the variational approximation.
\vspace{-0.4cm}

\paragraph{Expected number of inducing points :} Since the selection process is directly depending on the data, it is impossible to give an arbitrary bound.
However by adding assumptions on the distribution of $x$ one can 
\begin{theorem}
	Given a dataset i.i.d and uniformly distributed, i.e. $x \sim \mathcal{U}(0,a)^D$, and a SE kernel with lengthscale $l^D \ll 1$, the expected number of selected inducing points $M$ after parsing $N$ points is 
	\begin{align}
	\expec{}{M|N} \leq \frac{a^D-(a^D-\alpha)^{N+1}}{\alpha},
	\end{align}
	where $\alpha = \left(\frac{l\sqrt{-D\log \rho}}{2}\right)^D$.
	\label{thm:boundpoints}
\end{theorem}
The proof is given in the appendix \ref{appendix:proofnumind}.
As $N\rightarrow \infty$, this bound will converge to $a^D/\alpha$ which is the estimated number of overlapping hyper-spheres of radius $l\sqrt{-D\log \rho_{in}}$ to fill a hypercube of dimension $D$ with side length $a$.
This can be used as an upper bound for any data lying in a compact domain.
This confirms the intuition that the number of selected inducing points will grow faster with larger dimensions and a larger $\rho$ and with smaller lengthscales.

\paragraph{Expected performance on regression :} \citet{burt2019rates} derived a convergence bound for the inducing points approach of \cite{titsias2009variational}.
Even if they show this bound in an offline setting, their bound is still relevant for online problems.
They show that when $\setZ$ is sampled via a \kDPP~ process \cite{kulesza2011k}, i.e. a determinantal point process conditioned on a fixed set size, the difference between the ELBO and the log evidence $\log p (\mathcal{D})$ is bounded by
\vspace{-0.4cm}
\begin{align}
\expec{Z}{\|K_{X}-Q_{X}\|} \leq (M+1)\sum_{i=M+1}^N \lambda_i(K_{X})\label{eq:burt}
\end{align}\vspace{-0.7cm}

where $\lambda_i(K_{X})$ is the $i$-th largest eigenvalue of $K_{X}$ and $Q_{X}=K_{XZ}K_{Z}^{-1}K_{ZX}$ is the Nystr\"om approximation of $K_{X}$.

We derive a similar bound when using our algorithm instead of \kDPP sampling:
\begin{theorem}
	Let $\setZ$ be the set of inducing points locations of size $M$ selected via Algorithm \ref{alg:add} on the dataset $X$ of size $N$.
	\begin{align}
	\|K_{X}-Q_{X}\| \leq (N-M)\left(1-\frac{\rho^2}{1+ M(M-1)\rho}\right) \label{eq:expecbound}
	\end{align}
	where $K_X$ is the kernel matrix on $X$ and $Q_X$ is the Nystr\"om approximation of $K_X$ using the subset $\setZ$
	\label{thm:elbo}
\end{theorem}

The proof and an empirical comparison are given in the appendix \ref{appendix:prooftrace}.
\vspace{-0.45cm}
\section{Experiments}
\label{sec:experiments}
\begin{figure}
	\includegraphics[width=\columnwidth]{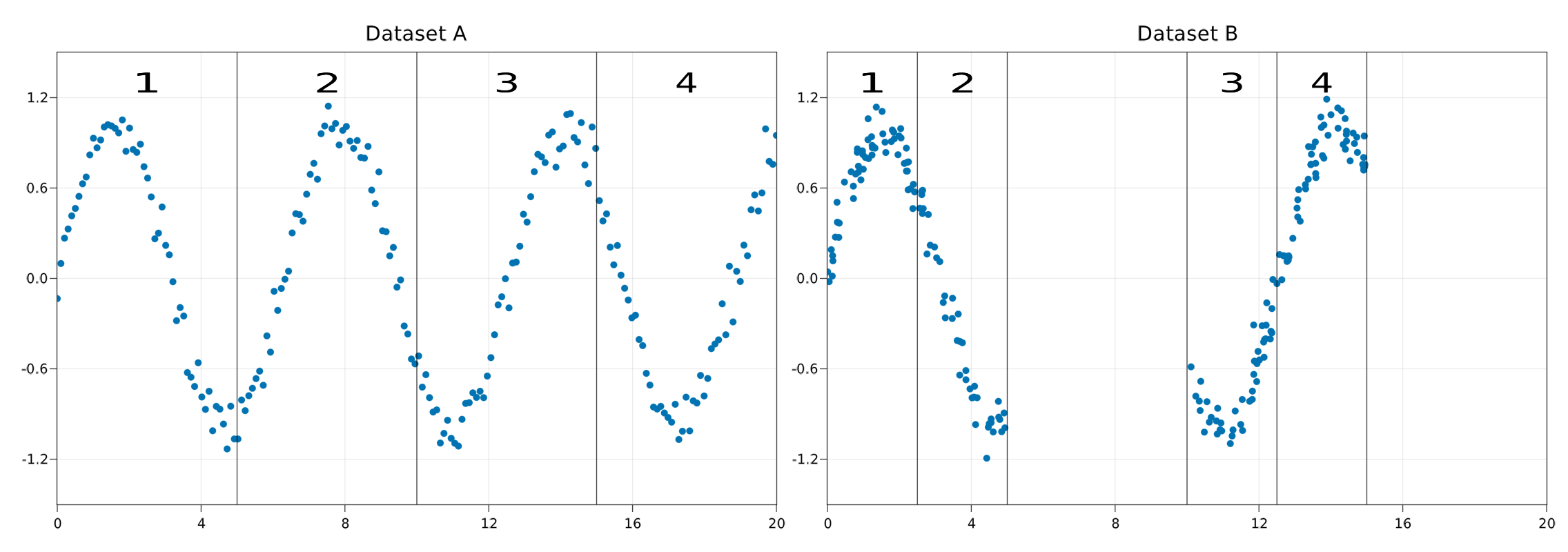}
	\includegraphics[width=\columnwidth]{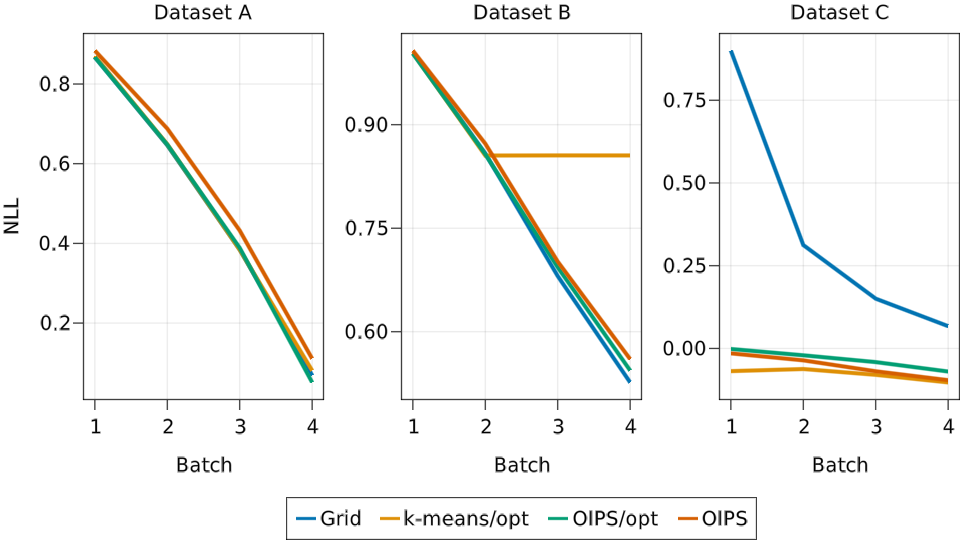}
	\caption{Toy datasets A and B, divived in 4 batches. Average Negative Test Log-Likelihood on a test set in function of number of batches seen. In a uniform streaming setting all methods perform similarly but having a gap blocks the convergence of a simple position optimization whereas in a non-compact situation the adaptive grid suffers in performance.}
	\label{fig:convergence}
\end{figure}

In this section we get a quick look on how our algorithm performs in different settings compared to approaches described in section \ref{sec:othermethods}.
We compare the online model \textbf{SSVGP} described in section \ref{sec:background} with different IP selection techniques. 
We select from the first batch via k-means and then optimize them (\textbf{k-means/opt}), select them via our algorithm and optimize them (\textbf{OIPS/opt}), select them via our algorithm but don't optimize them (\textbf{OIPS}) and finally create a \textbf{Grid} that we adapt according to new bounds.
We consider 3 different toy datasets, from which two are displayed in figure \ref{fig:convergence}.
The dataset A is a uniform time series and the output function is a noisy sinus.
The dataset B is an irregular time-series, with a gap in the inputs. The output function is also a noisy sinus.
Dataset C inputs are random samples from an isotropic multivariate 3D Gaussian and the output function is given by $\sin(||x||)/||x||$.
All datasets contain 200 training points and 200 test points.
For all experiments we use an isotropic SE kernel with fixed parameters.
For datasets A and B, \textbf{Grid} and \textbf{$k$-means} has 25 IPs while \textbf{OIPS} converged to around 20 IPs.
For dataset C, \textbf{Grid} has $10^3$ IPs, \textbf{$k$-means} 50, and both \textbf{OIPS} converged to 10 IPs
Figure \ref{fig:convergence} shows the evolution on the average negative log likelihood on test data after every batch has been seen.
On a uniform time-series context all methods are pretty much equivalent.
The presence of a gap, blocks the optimization of IP locations and impede inference of future points.
Whereas the grid suffers from being in high-dimensions and 
All details on the datasets, different training methods, hyper-parameters and optimization parameters used are to be found in appendix \ref{appendix:exp}.

\label{exp:performance}
\vspace{-0.5cm}
\section{Conclusion}
\label{sec:conclusion}
We presented a new algorithm, \namealg, able to select inducing points automatically for a GP in an online setting.
The theoretical bounds derived outperforms the previous work based on DPPs.
There is yet to improve the selection process to make it robust to outliers and to variations of the hyper-parameters.
Using for instance a threshold on the median or a mean on the $k$-nearest IPs could help to avoid picking adversarial points such as outliers.
We have only considered regression but our algorithm is also compatible with non-conjugate likelihoods.
Using augmentations approaches \cite{wenzel2019efficient,galy2019multi}, same performance can be attained.
Finally the most interesting improvement would be to use a non-stationary kernel \cite{remes2017non} and be able to automatically adapt the number of inducing points across the dataset.
\bibliographystyle{../icml2020}
\bibliography{cl_workshop_onlinegp.bib}

\newpage
\input{appendix}

\end{document}

%% file: commands.tex
\usepackage[usenames,dvipsnames]{xcolor}
\usepackage{amsthm}
\usepackage{mathtools}

\newcommand{\half}{\frac{1}{2}}

\newcommand{\logdet}[1]{\log\left|#1\right|}
\newcommand{\trace}{\text{tr}}

\newcommand{\KL}{\mathrm{KL}} 

\newcommand{\boldf}{\boldsymbol{f}}
\newcommand{\boldy}{\boldsymbol{y}}
\newcommand{\by}{\boldsymbol{y}}

\newcommand{\boldu}{\boldsymbol{u}}
\newcommand{\boldm}{\boldsymbol{m}}

\newcommand{\bmu}{\boldsymbol{\mu}}

\newcommand{\expec}[2]{\mathbb{E}_{#1}\left[#2\right]}

\newtheorem*{prop*}{Proposition}
\newtheorem{theorem}{Theorem}
\newtheorem*{theorem*}{Theorem}
\newtheorem*{lemma*}{Lemma}
\newtheorem*{corrolary*}{Corrolary}

\newtheorem*{definition*}{Definition}

%% file: appendix.tex
\appendix
\section{Derivations online GPs}
\label{appendix:elbo}
\subsection{ELBO}
Following \citet{bui2017streaming}, the ELBO for variational inference is defined as : 
\begin{align*}
\mathcal{L} =& -\KL\left(q_{t}(\boldu_t)||p(\boldu_t|\theta_{t})\right) + \expec{q_t(\boldu_t,\boldf_t)}{\log p(y_{t}|\boldf_t)}\\
& - \KL(q_{t}(\boldu_{t-1})||q_{t-1}(\boldu_{t-1})) \\
&+ \KL(q_{t}(\boldu_{t-1})||p(\boldu_{t-1}|\theta_{t-1}))
\end{align*}
The terms of the first line correspond to a classical SVGP problem and the second line express the KL divergence with the previous variational posterior.
The distributions are defined as :
\begin{align*}
q_{t}(\boldu_t) =& \mathcal{N}\left(\bmu_t,\Sigma_t\right)\\
p(\boldu_t|\theta_t) =& \mathcal{N}\left(0,K_{Z_t}\right)\\
q_{t}(\boldu_{t-1}) =& \int p(\boldu_{t-1}|\boldu_{t})q_t(\boldu_{t})d\boldu_t \\
=& \mathcal{N}\left(\kappa_{Z_{t-1}Z_{t}}\bmu_t,\widetilde{K}_{Z_{t-1}}\right)\\
\widetilde{K}_{Z_{t-1}} =& K_{Z_{t-1}}+\kappa_{Z_{t-1}Z_{t}}\Sigma_t \kappa_{Z_{t-1}Z_{t}}^{\top}\\
&-K_{Z_{t-1}Z_{t}}K^{-1}_{Z_{t}}K_{Z_{t}Z_{t-1}}\\
q_{t-1}(\boldu_{t-1}) =& \mathcal{N}\left(\bmu_{t-1},\Sigma_{t-1}\right)\\
p(\boldu_{t-1}||\theta_{t-1}) =& \mathcal{N}(0,\underbrace{K'_{Z_{t-1}}}_{\text{Given }\theta_{t-1}})
\end{align*}
The first terms ares
\begin{align*}
&\KL(q_t(\boldu_{t})||p(\boldu_{t}|\theta_t) =\\
&\quad\half\left(\logdet{K_{Z_t}}-\logdet{\Sigma_t}-M_t\right.\\
&\quad+\left.\trace(K_{Z_t}^{-1}\Sigma_t)+
 \bmu_t^\top K^{-1}_{Z_t}\bmu_t\right)
\end{align*}
And for $p(\by_t|\boldf_t)=\prod_{i=1}^B\mathcal{N}(y_i|f_i,\sigma)$. The expected log-likelihood is given by L
\begin{align*}
&\expec{q_t(\boldu_t,\boldf_t)}{\log p(\by_{t}|\boldf_t)}=-\frac{B}{2}\log 2\pi\sigma^2 \\
&\quad - \frac{1}{2\sigma^2}\sum_{i=1}^B (y_i-\kappa_{X_iZ_t}\mu_t)^2 + \widetilde{K} + \kappa_{X_iZ_t}\Sigma_t\kappa_{X_iZ_t}^\top
\end{align*}
Writing the second terms fully we get :
\begin{align*}
&\KL(q_{t}(\boldu_{t-1})||p(\boldu_{t-1}|\theta_{t-1})) =\\
& \quad \frac{1}{2}\left(\log |K'_{Z_{t-1}}| -\log|\widetilde{K}_{t-1}| -M_{t-1} \right.\\
& \quad + \trace((K'_{Z_{t-1}})^{-1}\widetilde{K}_{Z_{t-1}}) \\
&\quad \left. + (\kappa_{Z_{t-1}Z_{t}}\bmu_t)^\top(K'_{Z_{t-1}})^{-1}\kappa_{Z_tZ_{t-1}}\bmu_t\right)\\
&\KL(q_{t}(\boldu_{t-1})||q_{t-1}(\boldu_{t-1})) =\\ 
&\quad \frac{1}{2}\left(\log |\Sigma_{t-1}| -\log|\widetilde{K}_{Z_{t-1}}| -M_{t-1} \right.\\
& \quad + \trace(\Sigma_{t-1}^{-1}\widetilde{K}_{Z_{t-1}})\\
&\quad \left. + (\bmu_{t-1}-\kappa_{Z_{t}Z_{t-1}}\bmu_t)^\top \Sigma_{t-1}^{-1} (\bmu_{t-1}-\kappa_{Z_tZ_{t-1}}\bmu_t)\right)
\end{align*}

Subtracting the second term to the first we get:
\begin{align*}
&\KL_{t:t-1} =\\
&\quad  \KL(q_{t}(\boldu_{t-1})||p(\boldu_{t-1}|\theta_{t-1}))- \KL(q_{t}(\boldu_t)||q_{t-1}(\boldu_{t-1}))\\
& =\frac{1}{2}\left(\log|K'_{Z_{t-1}}|- \log |\Sigma_{t-1}| -\trace((\Sigma_{t-1}^{-1}-(K'_{Z_{t-1}})^{-1})\widetilde{K}_{Z_{t-1}}) \right.\\
&\quad - \bmu_{t-1}^\top \Sigma_{t-1}^{-1} \bmu_{t-1} + 2\bmu_{t-1} \Sigma_{t-1}^{-1} \kappa_{Z_{t-1}Z_{t}}\bmu_t \\
&\quad \left. - (\kappa_{Z_{t-1}Z_{t}}\bmu_t)^\top(\Sigma_{t-1}^{-1}-(K'_{Z_{t-1}})^{-1})(\kappa_{Z_{t-1}Z_{t}}\bmu_t)\right)\\
&= \frac{1}{2}\left(\log|K'_{Z_{t-1}}|- \log |\Sigma_{t-1}| - \trace(D_{t-1}^{-1}\widetilde{K}_{t-1})\right.\\
&\quad - \bmu_{t-1}^\top\Sigma_{t-1}^{-1}\bmu_{t-1} + 2\bmu_{t-1}\Sigma_{t-1}^{-1}\kappa_{Z_{t-1}Z_{t}}\bmu_t \\
&\quad \left.- (\kappa_{Z_{t-1}Z_{t}}\bmu_t)^\top D_{t-1}^{-1}(\kappa_{Z_{t-1}Z_{t}}\bmu_t)\right)\\
\end{align*}
Where $D_t = \left(\Sigma_t^{-1}-K_{Z_t}^{-1}\right)^{-1}$.\\

Taking the derivative of $\mathcal{L}$ given $\bmu_t$ and $\Sigma_t$ gives us directly the optimal solution for Gaussian regression:
\begin{align*}
\Sigma^*_t =& \left(\sigma^{-2}\kappa_{X_tZ_t}^\top\kappa^{\phantom{\top}}_{X_tZ_t}+\kappa_{Z_{t-1}Z_{t}}^\top D_{t-1}^{-1}\kappa^{\phantom{\top}}_{Z_{t-1}Z_{t}} + K_{Z_t}^{-1}\right)^{-1}\\
\bmu^*_t =& \Sigma_t\left(\kappa_{X_tZ_t}^\top\sigma^{-2}\boldy_t+ \kappa_{Z_{t-1}Z_{t}}^\top \Sigma_{t-1}\bmu_{t-1} \right)
\end{align*}

Rewritten in natural parameters terms:
\begin{align*}
\eta_1^{t} =& \kappa_{X_tZ_t}^\top \sigma^{-2} \boldy_t + \kappa_{Z_{t-1}Z_t}^\top \eta_1^{t-1}\\
\eta_2^{t} =& -\frac{1}{2}\left(\kappa_{X_{t}Z_{t}}^\top \sigma^{-2}I \kappa_{X_{t}Z_{t}}\right.\\ &\left. + \kappa_{Z_{t-1}Z_{t}}^\top \left(-2\eta_2^{t-1} - K_{Z_{t-1}}^{-1}\right) \kappa_{Z_{t-1}Z_{t}} + K_{Z_{t}}^{-1}\right)
\end{align*}

\subsection{Hyper-parameter derivatives}
Given $\theta$ a kernel hyperparameter and $J_{\Box\Box} = \frac{dK_{\Box\Box}}{d\theta}$ the derivatives are given by:
\begin{align*}
\frac{d\KL_{t:t-1}}{d\theta_t} =& -\frac{1}{2}\trace\left(D_{t-1}^{-1}\frac{d\widetilde{K}_{Z_{t-1}}}{d\theta_t}\right)\\ &+\bmu_{t-1}\Sigma_{t-1}^{-1}\frac{d\kappa_{Z_{t-1}Z_{t}}}{d\theta_t}\bmu_{t}\\
&- (\kappa_{Z_{t-1}Z_{t}}\bmu_{t})^\top D_{t-1}^{-1}(\frac{d\kappa_{Z_{t-1}Z_{t}}}{d\theta_t}\bmu_{t})\\
\frac{d\kappa_{Z_{t-1}Z_{t}}}{d\theta_t} =&  \frac{dK_{Z_{t-1}Z_{t}}}{d\theta_t}K_{Z_{t}}^{-1} + K_{Z_tZ_{t-1}}\frac{dK_{Z_t}^{-1}}{d\theta_t} \\=& (J_{Z_tZ_{t-1}}-\kappa_{Z_tZ_{t-1}} J_{Z_t})K_{Z_t}^{-1} = \iota_{Z_{t-1}Z_{t}}\\
\frac{d\widetilde{K}_{Z_{t-1}}}{d\theta_t} =& \frac{dK_{Z_{t-1}}}{d\theta_t} + 2\frac{d\kappa_{Z_{t-1}Z_{t}}}{d\theta_t}\Sigma_t\kappa_{Z_tZ_{t-1}}^\top \\
&- \frac{d\kappa_{Z_{t-1}Z_{t}}}{d\theta_t}K_{Z_{t}Z_{t-1}} - \kappa_{Z_{t-1}Z_{t}}\frac{dK_{Z_{t}Z_{t-1}}}{d\theta_t}\\
=&J_{Z_{t-1}} + 2\iota_{Z_{t-1}Z_{t}} \Sigma_t \kappa_{Z_{t-1}Z_{t}}^\top \\
&- \iota_{Z_{t-1}Z_{t}} K_{Z_{t}Z_{t-1}} - \kappa_{Z_{t-1}Z_{t}}J_{Z_{t}Z_{t-1}}
\end{align*}
\begin{align*}
\frac{d\KL(q_t(\boldu_{t})||p(\boldu_{t}|\theta_t)}{d\theta_t}
\end{align*}
Special derivative given the variance : 
\begin{align*}
\frac{dKL_a}{dv} = & - \frac{1}{2}\left(\trace\left(D_a^{-1}\left[\frac{1}{v}(K_{aa}-K_{ab}K_{bb}^{-1}K_{ba})\right]\right)\right)
\end{align*}

\subsection{Comparison with SVI}

If we take the special case where inducing points do not change between iterations, then $\kappa_{Z_{t-1}Z_{t}} = I$ and ${K_{Z_{t-1}} = K_{Z_t}}$. The updates become
\begin{align*}
\eta_1^{t} =& \kappa_{X_tZ_t}^\top \sigma^{-2} \boldy_t + \eta_1^{t-1}\\
\eta_2^{t} =& -\frac{1}{2}\left(\kappa_{X_tZ_t}^\top \sigma^{-2} \kappa_{X_tZ_t} + \left(-2\eta_2^{t-1} - K_{Z_t}^{-1}\right) + K_{Z_t}^{-1}\right) \\
=& -\frac{1}{2}\kappa_{X_{t}Z_t}^\top \sigma^{-2} \kappa_{X_{t}Z_t} + \eta_2^{t-1}
\end{align*}
Compared to the SVI updates: 
\begin{align*}
\eta_1^{t} = & \eta_1^{t-1} + \rho \left(\frac{N}{|B|}\left(\kappa_{X_tZ_t}^\top \sigma^{-2} \boldy_t\right) - \eta_1^{t-1} \right)\\
\eta_2^{t} = & \eta_2^{t-1} + \rho \left(-\frac{1}{2}\left(\frac{N}{|B|}\kappa_{X_tZ_t}^\top \sigma^{-2} \kappa_{X_tZ_t} + K_{Z_t}^{-1}\right)-\eta_2^{t-1}\right)
\end{align*}
If we ignore $\rho$ by setting it as 1:
\begin{align*}
\eta_1^{t} = & \frac{N}{|B|}\left(\kappa_{X_tZ_t}^\top \sigma^{-2} \boldy_t\right)\\
\eta_2^{t} = & -\frac{1}{2}\left(\frac{N}{|B|}\kappa_{X_tZ_t}^\top \sigma^{-2} \kappa_{X_tZ_t}+ K_{Z_t}^{-1}\right)
\end{align*}
We forget completely the previous $\eta_1$. \\
To make it directly comparable to streaming:
\begin{align*}
\intertext{SVI}
\eta_1^{t+1} = & (1-\rho)\eta_1^{t} + \rho \left(\frac{N}{|B|}\left(\kappa_f^\top \sigma^{-2} y\right) \right)\\
\eta_2^{t+1} = & (1-\rho)\eta_2^{t} + -\frac{1}{2} \rho \left(\frac{N}{|B|}\kappa_f^\top \sigma^{-2} \kappa_f + K_{bb}^{-1}\right)\\
\eta_1^t = & (1-\rho)^t\eta_0 + \sum_{i=1}^t (1-\rho)^{i-1}\rho \frac{N}{|B|}\kappa_f^\top \sigma^{-2} y^i\\
\intertext{Streaming}
\eta_1^{t+1} =& \eta_1^{t} + \kappa_f^\top \sigma^{-2} y\\
\eta_2^{t+1} =& \eta_2^{t}-\frac{1}{2}\kappa_f^\top \sigma^{-2} \kappa_f
\end{align*}

\section{Deterministic algorithm as a DPP half-greedy sampling}
\label{appendix:pdfdpp}
\begin{figure}
	\centering
	\includegraphics[width=\columnwidth]{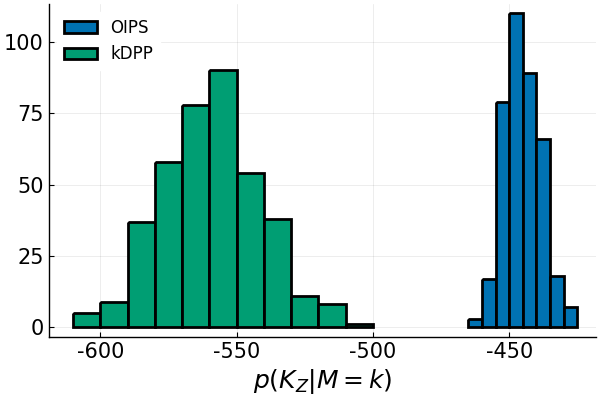}
	\caption{Histogram of $p(Z|k=M)$ for the OIPS algorithm and \kDPP sampling}
	\label{fig:pmfdpp}
\end{figure}

We proceed to a simple experiment, where given a dataset, Abalone ($N=4177,D=7$), we repeatedly shuffle the data.
We apply algorithm \ref{alg:add} parsing all the data to get the subset $Z_{OIPS}$.
We use the resulting number of inducing points $k$ as a parameter to sample from a \kDPP~and obtain $Z_{kDPP}$. 
We compute the probabilities of $\log p(Z_{OIPS}|M=k)$ and $\log p(Z_{kDPP}|M=k)$ and report the histogram of the probabilities on figure \ref{fig:pmfdpp}
One can observe that the probability given by the OIPS algorithm is consistently higher as well as more narrow then the sampling.
This can be explained by the fact that we deterministically constrain all the points to have a certain distance from each other and therefore put a deterministic limit on the determinant of $K_Z$.
\section{Proof Theorem \ref{thm:boundpoints} : Bound on the number of points}
\label{appendix:proofnumind}
Algorithm \ref{alg:add} can be interpreted as filling a domain with closed balls, where balls intersections are allowed but no center can be inside another ball.
For a SE kernel we can compute the radius $r$ (in euclidean space) of these balls :
\begin{align*}
k(x,x') &= \rho_{in}\\
\exp\left(-\frac{||x-x'||^2}{h^2}\right) &=\rho_{in}\\
||x-x'||^2 &= -h^2\log \rho_{in}\\
r &= h\sqrt{-\log \rho_{in}}
\end{align*}

We can bound the volume of the union of the balls by the union of inscribed hypercubes.
The length of an inscribed hypercube in an hypersphere of radius $r$ is $l = r\sqrt{D}/2$. 
Since the volume of the hypercube is defined to be smaller, this gives us an upper bound on the expected number of inducing points.
Defining as $K_n$ the number of inducing points at time $n$, the probability of having a point outside of the union of all $k$ hypercubes is
\begin{align*}
p(K_{n+1}=k+1|K_n=k) =& \max\left(a^D-\sum_{i=1}^kl^D\right)\\
=& \max\left(a^D-kl^D,0\right)\\
p_k^+=& \max\left(a^D-k\alpha,0\right)\\
\intertext{Where $\alpha = \left(\frac{r\sqrt{D}}{2}\right)^D$, is the volume of one hypercube and therefore the probability of a new sample to appear in it. }
\intertext{The probability of keeping the same number of points is}
p(K_{n+1}=k|K_n=k) =& \min\left(\sum_{i=1}^k l^D,1\right) \\
p_k^==&  \min(k\alpha,1)
\end{align*}

We now consider the problem as a Markov chain where the state $p$ is represented by a vector $\{p_i\}_{i=1}^N$ where $p_i=1$ if there are $i$ inducing points. The transition matrix $P$ is given by : 
\begin{align*}
P = \left(
\begin{array}{cccc}
p_1^= & 0 & 0 & 0\\
p_1^+ & p_2^= & 0 & 0\\
0 & p_2^+ & \ddots & 0 \\
0 & 0 & \ddots & 0\\
0 & 0 & p_{N-1}^+& p_N^=
\end{array}\right)
\end{align*}
If we define that we start with inducing points the initial state is $p^1 =\{1,0,\ldots,0\}^\top$, the probability of having $k$ balls after $n$ steps is $p(K_n=k|p^1) = \left(P^{n}p^1\right)_k$ while the expected number of pointsis given by ${\sum_k k\cdot p(K_n=k|p^1)}$.



These sequence can be complex to compute.
Instead we can approximate the final expectation by recursively computing the update given the expectation at the previous step:
\begin{align*}
\expec{p(K_{n+1}|K_{n}=\expec{}{K_n})}{K_{n+1}}\\
= \expec{}{K_n} \expec{}{K_n}\alpha + (\expec{}{K_n}+1)(a^D-\expec{}{K_n}\alpha)\\
= a^D\expec{}{K_n} + a^D - \expec{}{K_n}\alpha = a^D+\expec{}{K_n}(a^D-\alpha)
\end{align*}
This is an arithmetico-geometric suite and given the original condition $\expec{}{K_0}=1$ and since $\alpha < a^D$ we can get a closed form solution for $\expec{}{K_n}$:
\begin{align*}
\expec{}{K_n} =&(a^D-\alpha)^n\left(1 - \frac{a^D}{\alpha}\right) + \frac{a^D}{\alpha}\\
=& \frac{a^D-(a^D-\alpha)^{n+1}}{\alpha}
\end{align*}

\subsection{Empirical Comparison}
\begin{figure}[H]
	\includegraphics[width=\columnwidth]{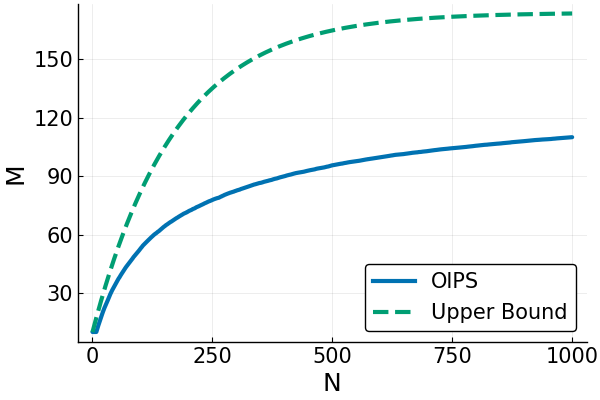}
	\caption{Bound on the number of inducing points accepted $M$ given the number of seen points $N$ vs the empirical estimation}
	\label{fig:boundM}
\end{figure}

We show the realization of this bound on uniform data with 3 dimensions, $\rho=0.7$ and $l = 0.3$ on figure \ref{fig:boundM}. 

\section{Proof theorem \ref{thm:elbo} : Bounding the ELBO}
\label{appendix:prooftrace}
We follow the approach of \citet{burt2019rates} and \citet{belabbas2009spectral}.
\citet{burt2019rates} showed that the error between the ELBO and the log evidence was bounded by $\|K_{X} - K_{XZ}K_{Z}^{-1}K_{ZX}\|$.
Where $\|\cdot\|$ is the Froebius norm.
Using a k-DPP sampling \cite{kulesza2011k}, they were able to show a bound on the expectation of this norm.
We follow similar calculations with our deterministic algorithm for fixed kernel parameters.
Let be $K_{X}$ the kernel matrix of the full dataset and $K_{Z}$ the submatrix given the set of points $\{Z_i\}_{i=1}^M$. 
The Schur complement of $K_{ZZ}$, $S_C(K_{ZZ})$ in $K_{XX}$ is given by $K_{X} - K_{XZ}K_{Z}^{-1}K_{ZX}$.
Following a similar approach then \citet{belabbas2009spectral} we bound the norm by the trace:
\begin{align*}
\|S_C(K_{ZZ})\| = \sqrt{\sum_{j=1}^{N-M} \overline{\lambda}_j} \leq \sum_{j=1}^{N-M} \overline{\lambda}_j = \trace(S_C(K_{ZZ}))
\end{align*}
Using the definiton of $S_C(K_{ZZ})$ we get :
\begin{align*}
\trace(S_C(K_{ZZ})) = \sum_{i=1}^{N-M} K_{X_i}-K_{X_iZ}K_{Z}^{-1}K_{ZX_i}\\
\end{align*}
where every element of the sum is a scalar.
Taking $W^\top\overline{\Lambda} W$ the eigendecomposition of $K_{Z}^{-1}$, $w_i = WK_{X_iZ}$  and assuming a kernel variance $v$ of 1 (although generalizable to all variances) and a translation invariant kernel such that $k(x,x)=1$ we get :
\begin{align*}
&K_{X_i}-K_{X_iZ}K_{Z}^{-1}K_{ZX_i}=1-w_i^\top\Lambda w_i = 1 - \sum_{j=1}^M \overline{\lambda_j}(w_i)_j^2\\
&\leq 1 - \overline{\lambda}_{\min} \|w_i\|^2 = 1-\overline{\lambda}_{\min}\|K_{X_iZ}\|^2\leq 1- \overline{\lambda}_{\min}\rho^2
\end{align*}
Where we used the fact that at least $X_i$ was close enough to at least one $Z_j$ such that $k(X_i,Z_j)>\rho$.
For clarity we replace $\overline{\lambda}_{\min}=\lambda^{-1}_{\max}$ where $\lambda_{\max}$ is the largest eigenvalue of $K_Z$.
When summing over the trace we get the final bound :
\begin{align*}
\|K_X-K_{XZ}K_Z^{-1}K_{ZX}\| \leq (N-M)\left(1-\frac{\rho^2}{\lambda_{\max}}\right)
\end{align*}

Now by construction all off-diagonal terms of $K_{Z}$ are smaller than $\rho$.
Using the equality \cite{Stewart90}
\begin{align*}
|\lambda_i(A)-\lambda_i(B)| \leq \|A-B\|,\quad \forall i=1,\ldots,N
\end{align*}
We get that 
\begin{align*}
|\lambda_{\max}(K_Z) - 1| \leq& \|K_{Z}-I\|_2 = \sqrt{\sum_{i\neq j}\left(K_{Z}\right)^2_{ij}}\\
\leq& M(M-1)\rho\
\end{align*}
Assuming $\lambda_{\max}(K_Z)\geq 1$, we get
\begin{align*}
\lambda_{\max}(K_Z) \leq 1 + M(M-1)\rho_{out}
\end{align*}
Getting then the final bound :
\begin{align*}
\|K_X-Q_X\| \leq (N-M)\left(1-\frac{\rho^2}{1+M(M-1)\rho}\right)
\end{align*}

\subsection{Empirical Comparison}

\begin{figure}
	\includegraphics[width=\columnwidth]{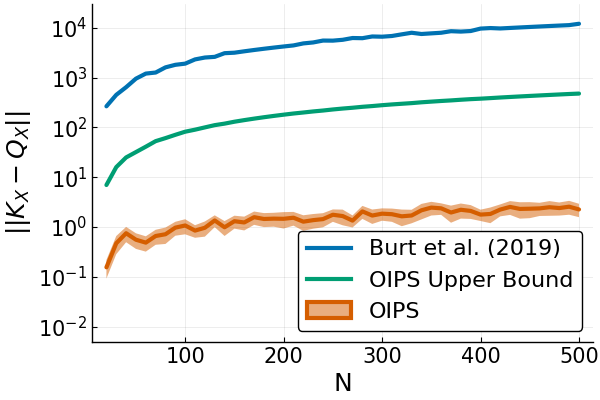}
	\caption{Evaluation of the $\|K_X-Q_X\|$ given the OIPS algorithm and computation of the bound from \citet{burt2019rates} given in equation \ref{eq:burt} and our bound given in equation \ref{eq:expecbound}}
	\label{fig:expecbound}
\end{figure}

These bounds are difficult to compare due to the different parameters characterizing them.
Nevertheless we give an example by comparing the bound and the empirical value on toy data drawn uniformly in 3 dimensions in figure \ref{fig:expecbound}.
For each $N$ we ran our algorithm and input the required $M$ in the bounds as the resulting number of selected inducing points.
We show in the section \ref{sec:experiments} the empirical effect on the accuracy and on the number of points given the choice of $\rho$.

\section{Experiments parameters}
\label{appendix:exp}
For every problem we use an isotropic Squared Exponential Kernel : 
\begin{align*}
k(\mathbf{x},\mathbf{x}') = v\exp\left(-\frac{\|\mathbf{x}-\mathbf{x}'\|^2}{h^2}\right)
\end{align*}
Where $h$ is initialized by taking the median of the lower triangular part of the pairwise distance matrix of the first subset of points and fixed for the rest of the training.
Future work will involve working with kernel parameter optimization as well.
We fix the noise of the Gaussian likelihood to $\sigma^2 = 0.01$.

IPs were optimized via ADAM ($\alpha=10^{-2}$).